\documentclass{article}
\usepackage{ijcai22}

\usepackage{times}
\usepackage{soul}
\usepackage{url}
\usepackage[hidelinks]{hyperref}
\usepackage[utf8]{inputenc}
\usepackage[small]{caption}
\usepackage{graphicx}
\usepackage{amsmath}
\usepackage{amsthm}
\usepackage{booktabs}
\usepackage{algorithm}
\usepackage{algorithmic}
\usepackage{amssymb}
\usepackage{makecell}

\title{Multi-Target Active Object Tracking with Monte Carlo Tree Search \\ and Target Motion Modeling}

\author{
    Zheng Chen, Jian Zhao, Mingyu Yang, Wengang Zhou, Houqiang Li
}

\begin{document}

\maketitle

\begin{abstract}
In this work, we are dedicated to multi-target active object tracking (AOT), where there are multiple targets as well as multiple cameras in the environment. The goal is maximize the overall target coverage of all cameras. Previous work makes a strong assumption that each camera is fixed in a location and only allowed to rotate, which limits its application. In this work, we relax the setting by allowing all cameras to both move along the boundary lines and rotate. In our setting, the action space becomes much larger, which leads to much higher computational complexity to identify the optimal action. To this end, we propose to leverage the action selection from multi-agent reinforcement learning (MARL) network to prune the search tree of Monte Carlo Tree Search (MCTS) method, so as to find the optimal action more efficiently. Besides, we model the motion of the targets to predict the future position of the targets, which makes a better estimation of the future environment state in the MCTS process. We establish a multi-target 2D environment to simulate the sports games, and experimental results demonstrate that our method can effectively improve the target coverage.
\end{abstract}

\section{Introduction}
Traditional visual object tracking (VOT) task \cite{henriques2014high,nam2016learning} aims to estimate a bounding box of the interested target in a series of frames. In this task, the video data is pre-captured, and the tracking precision will be degraded when the target is occluded or moves beyond the field of view. To avoid this problem, AOT controls the adjustment of the camera according to the visual scene to ensure that the target is in the field of view. AOT has many applications in practical scenarios, such as unmanned aerial vehicles (UAV), intelligent robots, sports events, \emph{etc}.

In AOT, the targets to be tracked may be single or multiple, and the corresponding performance shall be evaluated with different metrics. For single-target AOT task, we evaluate the tracking performance by checking whether the target is tracked by the camera and the location of the target in the camera observation \cite{luo2018end}. However, in multi-target AOT task, since multiple targets are scattered throughout the environment, it is difficult to use conventional evaluation indicators to define the tracking performance. To this end, a new metric, \emph{i.e.}, target coverage \cite{guvensan2011coverage}, is used to evaluate the tracking performance, which is also reasonable in practical application. For example, in a football game, we expect that the cameras around the field can cover as many players as possible. In this work, we focus on the multi-target situation, and try to maximize the overall target coverage.

\begin{figure}[t]
\centering
\includegraphics[width=1.\linewidth]{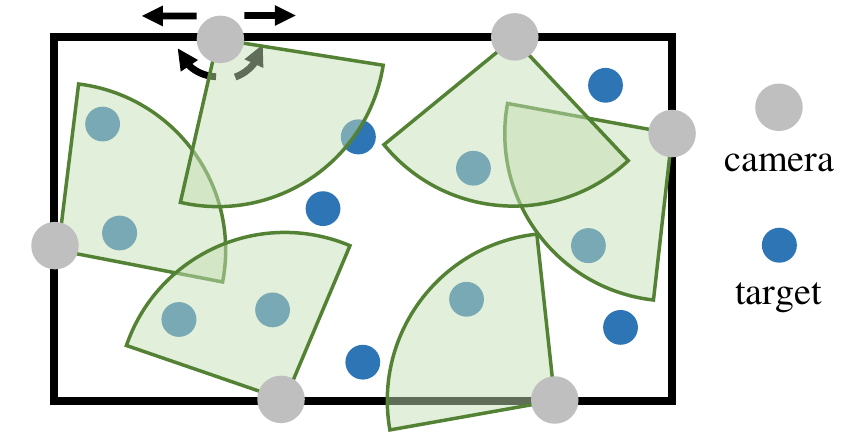}
\caption{Overview of the 2D environment. The black rectangle represents the range of the sports field. The blue points indicate the players (targets). The gray points denote the cameras. The green sectors refer to the range of the camera's field of view. The camera can move along the sports field and rotate.}
\label{env}
\end{figure}

Previous work \cite{li2018heading} with multiple targets finds the points that are not easy to be covered in the environment, and designs a cooperation mechanism to reduce the overlaps between cameras. In \cite{zhang2016local}, the ares is divided into different cells to transform the coverage problem into cell coverage problems. However, these methods assume the cameras are fixed in location. When the location of camera is variable, these traditional methods will encounter difficulties. In \cite{xu2020learning}, multi-agent reinforcement learning is firstly introduced to solve the multi-target AOT problem. It assigns the targets to each camera according to the position of the cameras and targets, and then control the camera to cover the assigned targets. However, this method uses the information of God's perspective, that is, it uses the position information of the target which is actually unavailable in the field of view of any camera.

Our work solves the shortcomings of previous work mentioned above, and proposes a Monte Carlo tree search and target motion modeling method to improve the target coverage. Since targets do not take Brownian motion in the real-word situation, \emph{i.e.}, the motion of targets in the adjacent time period has a strong correlation. Inspired by this, we predict the target position according to the previous information, then we use MCTS to search future states of the environment, in order to find the best action of the cameras. Since the joint action space of multiple cameras is large, we use the action output of the multi-agent reinforcement learning network to prune the search tree, which greatly improves the efficiency of tree search. In our work, the input of MARL network contains the target information in the field of view of all cameras and the position information of cameras, which is called centralized information. In addition, we allow the cameras to move in location which can not be achieved in traditional methods.

%The output of MARL network contains the joint actions of cameras, and the actions of each camera include the movement of location and rotation. Individual reward and team reward are used simultaneously in MARL network training, so that the camera not only considers the targets within its own field of vision, but also takes into account the overall target coverage when selecting actions. In addition, since targets do not take Brownian motion in the real-word situation, \emph{i.e.}, the motion of targets in the adjacent time period has a strong correlation. Therefore, we can roughly predict the positions of targets according to the previous information. Inspired by this, we use MCTS method to search future states of the environment, in order to find a better action for the cameras. %The output of MARL network is used to prune the search tree of MCTS method.

To verify the effectiveness of our method, we build a 2D environment to simulate the scenes of sports games (\emph{e.g.}, football, basketball, and volleyball). In our environment, cameras can obtain their posture and the coordinates of targets in the field of view. Besides, we control the location and rotation of the camera to maximize the target coverage. The schematic diagram of the environment is shown in Figure
\ref{env}.

In summary, our main contributions are three-fold:

\begin{itemize}
	\item We generalize the multi-target AOT problem by increasing the cameras' degrees of freedom in multiple targets AOT. Such a camera setting is also closer to the scenario of live sports field pickup in the real-world situation.
	\item We propose a MARL network to predict the action of cameras in a centralized setting, which makes full use of the information between cameras.
	\item We model the regularity of target motion to predict the future state of the environment, and integrate it with MCTS method to obtain a better action based on future information.
\end{itemize}

\section{Related Work}
In this section, we briefly review the recent works on active object tracking from two different aspects: single-target and multi-target.
\subsection{Single-target Active Object Tracking}
The development of deep reinforcement learning \cite{mnih2013playing,hessel2018rainbow} provides a foundation for active object tracking tasks. Different from supervised learning, reinforcement learning doesn't need labelled data, and only rewards are used to describe the quality of action in a state. Thus reinforcement learning is mainly committed to finding the optimal policy to maximize the cumulative reward, which is produced in the interaction between agent and environment. %Early work on reinforcement learning is focused on a single agent, and is usually formulated as Markov Decision Process (MDP).

The first active object tracking work \cite{luo2018end} trains deep reinforcement learning network through end-to-end method. It takes the raw frames of the camera as the input of the network, and outputs the discrete action of the camera through the processing of convolutional layer and LSTM. It adopts two types of virtual environments ViZDoom \cite{kempka2016vizdoom} and Unreal Engine with UnrealCV \cite{qiu2017unrealcv} for simulated training and testing. The experiment uses accumulated reward (AR) and episode length (EL) metrics to show the great performance of the end-to-end method. In the follow-up work, \cite{zhong2018ad} adds the adversarial mechanism in the process of network training, which let the tracker and target train under a zero sum reward. The target no longer moves according to the preset rules, but tries to escape the tracker as far as possible. In the face of more challenging target, the final trained tracker has better robustness. \cite{self} considers the distractors in the environment. To address this issue, they introduce an attention mechanism to make the network focus on the target of interest. CMC-AOT \cite{li2020pose} considers the case that multiple cameras cooperate to track a single target in active object tracking. CMC-AOT builds vision-based controller and pose-based controller, and the cameras select different controllers to perform their actions according to whether they can observe the target or not. Different from the above works, our work is dedicated to solving the problem of multiple cameras and multiple targets.

%In recent years, multi-agent reinforcement learning has attracted substantial attention in many applications, such as multi-object tracking \cite{rosello2018multi}, traffic light control \cite{torabi2020deployment,ma2020feudal,singh2020hierarchical}, real-time game \cite{vinyals2019alphastar} and so on. Compared with the single agent reinforcement learning task, multi-agent reinforcement learning has a larger action space. The traditional actor-critic method usually uses a team reward to train each agent separately. However, this training strategy is unstable. When an agent takes the same action in a certain state, the team reward may be different because other agents perform different actions. In order to solve this problem, \cite{lowe2017multi} adds the information of other agents to the critic network. To evaluate the contribution of each agent to the environment, COMA \cite{foerster2018counterfactual} utilizes a counterfactual baseline to obtain the impact of each agent's action on the environment.

\subsection{Multi-target Active Object Tracking}
When there are multiple freely moving targets in the environment, it is impossible for one camera to track all targets at the same time. Therefore, accumulated reward and episode length are no longer applicable. At this time, we are more concerned about the target coverage. For the case of multiple cameras, we use multi-agent reinforcement learning instead of the original reinforcement learning. Recently, multi-agent reinforcement learning has attracted substantial attention in many applications, such as multi-object tracking \cite{rosello2018multi}, traffic light control \cite{torabi2020deployment,ma2020feudal,singh2020hierarchical}, real-time game \cite{vinyals2019alphastar} and so on.

Previous multi-target active object tracking work \cite{xu2020learning} proposes a novel HiT-MAC framework which maximizes target coverage while reducing the energy consumption. HiT-MAC first uses the coordinator to assign the targets to each camera, and then uses the executor to control each camera to track the assigned targets. However, in their setting, the locations of cameras are immutable, and the coordinator uses the location information of all targets, even those not in the field of view of any camera. Our work gives the camera freedom to move in location, and avoids using the target location information that is not within the visual range of the camera. At the same time, we further optimize the camera action by predicting the future location of the targets.

\begin{figure*}[t]
\centering
\includegraphics[width=.92\linewidth]{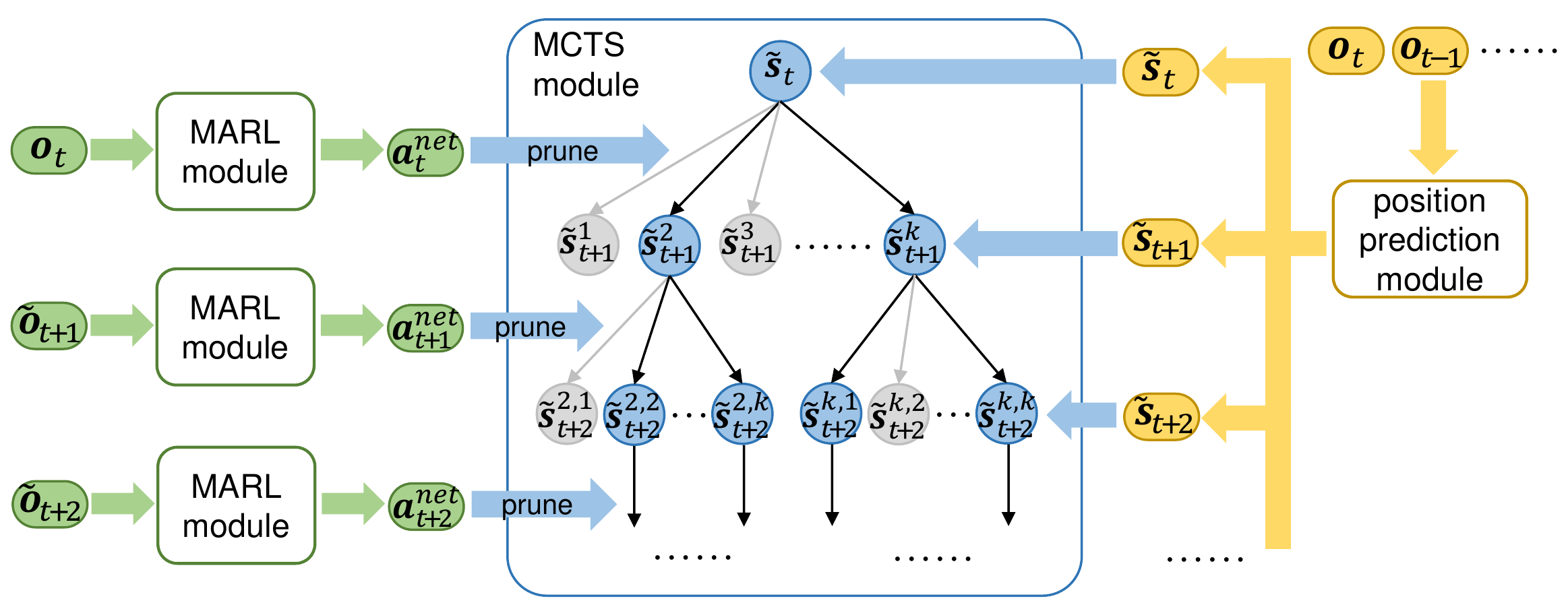}
\caption{The tracking pipeline of our method. Our method consists of MCTS module, position prediction module, and MARL module. $\boldsymbol{o}_t$ denotes the centralized observations. In time step $t$, position prediction module estimate the environment state $\boldsymbol{s}_{t}, \boldsymbol{s}_{t+1}, \cdots$. Based on the estimated future environment state, MCTS module finds the optimal action by searching the camera action. The output of MARL module $\boldsymbol{a}^{net}_t$ helps to remove unnecessary branches (gray nodes in the figure) in the search tree, so that we can obtain the final action $\boldsymbol{a}^{mcts}_t$ of cameras more efficiently.}
\label{figmcts}
\end{figure*}

\section{Method}
Our method includes three modules, \emph{i.e.,} Monte Carlo tree search (MCTS), position prediction, and multi-agent reinforcement learning (MARL). MCTS module is the main part of our method. It searches the camera action according to the current and future environment state, so as to obtain the optimal action of the camera. The position prediction module uses the current and history observation states to predict the future environment states, which provides a basis for Monte Carlo tree search. The action output by MARL module prunes the search tree, which improves the search efficiency. The relationship of the three modules is shown in Figure \ref{figmcts}. In the following, we first introduce the 2D environment, then discuss each module in detail.

\subsection{2D Environment}
In order to simulate the scenes of sports games, we set the field as a rectangle. As shown in Figure \ref{env}, the cameras are set at the boundary and allowed to move along the boundary lines and rotate. The moving range of targets is limited to the rectangle. We use $n$ and $m$ to represent the number of cameras and targets in the environment, respectively.

\paragraph{State Space}
The environment state $\boldsymbol{s}_t$ in time step $t$ includes the posture of all cameras and targets. The posture of the $i$-th camera $c_{t,i}$ can be represented as the tuple $c_{t,i}=(\alpha_{t,i}, x^c_{t,i}, y^c_{t,i})$, where $\alpha_{t,i}$ refers to the rotation of the $i$-th camera, and $(x^c_{t,i}, y^c_{t,i})$ indicates the location of the $i$-th camera. Similarly, the posture of the $j$-th target can be represented as the tuple $p_{t,j}=(x^p_{t,j}, y^p_{t,j})$. %where $x^p_{t,j}, y^p_{t,j}$ refers to the location of the $j$-th target.
\paragraph{Observation Space}
In time step $t$, the $i$-th camera's observation $\boldsymbol{o}_{t,i}$ consists of the posture of the $i$-th camera $c_{t,i}$ and the relative position between the $i$-th camera and all targets. The position relationship between the $i$-th camera and the $j$-th target can be represented as the tuple $cp_{t,ij}=(d_{t,ij}, \sin\theta_{t,ij}, \cos\theta_{t,ij})$, where $d_{t,ij}$ and $\theta_{t,ij}$ refer to the distance and angle between the $i$-th camera and the $j$-th target, respectively. In most cases, a  single camera will not observe all targets. For unobserved targets, the tuple $cp_{t,ij}$ will be filled with $-1$. The observation of the $i$-th camera can be written as $\boldsymbol{o}_{t,i}=(c_{t,i}, cp_{t,i1}, cp_{t,i2}, \cdots, cp_{t,im})$. The observations of all cameras are combined as the centralized observation $\boldsymbol{o}_{t}=(\boldsymbol{o}_{t,1},\boldsymbol{o}_{t,2},\cdots,\boldsymbol{o}_{t,n})$.
\paragraph{Action Space}
The joint action $\boldsymbol{a}_t$ in time step $t$ includes the action options $a_{t,i}$ of all cameras in the environment, \emph{i.e.}, $\boldsymbol{a}_t=(a_{t,1}, a_{t,2}, \cdots, a_{t,n})$. Each camera in the environment is constrained to move at the edge of the field and rotate. So there are 9 discrete action options for each camera.
%Since the location of movement is limited, we can only use one dimension to represent the movement of the camera. The camera location movement includes $Loc = \{Forward, Backward, Stay\}$, and the camera rotation movement includes $Rot = \{Clockwise, Anticlockwise, Stay\}$. Due to the location and rotation are independent, the action of the $i$-th camera $a_{t,i} \in Loc \times Rot$, which means each camera has 9 discrete action options.

\subsection{Position Prediction by Target Motion Modeling}
%When we use MCTS method to search for the optimal action $\boldsymbol{a}^{mcts}_{t}$, we need to know the future environment state $\boldsymbol{s}_{t}, \boldsymbol{s}_{t+1}, \cdots, \boldsymbol{s}_{t+D}$, where $D$ denotes the search depth of MCTS method. 
In position prediction module, we attempt to predict the current environment state $\boldsymbol{s}_{t}$ and the future environment state $\boldsymbol{s}_{t+1}, \boldsymbol{s}_{t+2}, \cdots, \boldsymbol{s}_{t+D}$, where $D$ denotes the search depth of MCTS module. However, in time step $t$, we can only obtain the centralized observation $\boldsymbol{o}_{t}$. First of all, we estimate $\boldsymbol{s}_{t}$ with $\boldsymbol{o}_{t}$.

\paragraph{Current State Estimation}
Since centralized observation $\boldsymbol{o}_{t}$ contains the posture of cameras $c_{t,i}$ and the relative position between target and camera $cp_{t,ij}$, for the target within the field of view, we calculate its position as follows,
\begin{equation}\label{st}
\left\{
\begin{aligned}
x^p_{t, j} & = x^c_{t,i} + d_{t,ij} \sin(\alpha_{t,i}+\theta_{t,ij}) \\
y^p_{t, j} & = y^c_{t,i} + d_{t,ij} \cos(\alpha_{t,i}+\theta_{t,ij})
\end{aligned}
\right.
.
\end{equation}
So far, the estimated $\tilde{\boldsymbol{s}}_t$ contains the position of all cameras and the target within the field of view. Similarly, we may estimate the historical state $\tilde{\boldsymbol{s}}_{t-1}$ according to $\boldsymbol{o}_{t-1}$.

\begin{figure}[t]
\centering
\includegraphics[width=0.95\linewidth]{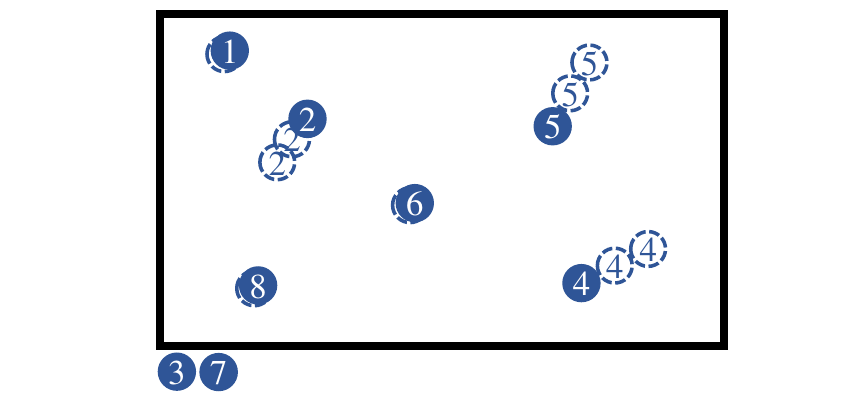}
\caption{Overview of our prediction method. The blue solid points denotes the predicted target position in time step $t+1$, and the hollow points with dotted border denotes the target position in previous time steps. Since some targets are not observed in previous time steps, we don't mark these targets in the figure, for example, target 3 and 7.}
\label{pre}
\end{figure}

\paragraph{Future State Estimation}
Due to the future centralized observations are not available, we use the $\tilde{\boldsymbol{s}}_{t-1}$ and $\tilde{\boldsymbol{s}}_t$ to estimate the $\tilde{\boldsymbol{s}}_{t+1}$.

We assume that the target moves in a uniform straight line in a short period of time. This assumption is
reasonable because the motion pattern of a target will have a great correlation with its motion pattern of the previous time. Therefore, if a target's position in time step $t$ is $(x^p_t, y^p_t)$ and its position in last time step $t-1$ is $(x^p_{t-1}, y^p_{t-1})$, then its position in time step $t+t_0$ can be estimated as,
\begin{equation}\label{positionpredict}
\left\{
\begin{aligned}
\tilde{x}^p_{t+t_0} & = x^p_t + t_0 (x^p_t-x^p_{t-1}) \\
\tilde{y}^p_{t+t_0} & = y^p_t + t_0 (y^p_t-y^p_{t-1})
\end{aligned}
\right.
.
\end{equation}
In Equation \ref{positionpredict}, $t_0$ should not be too large, because it only conforms to this equation in a short time period.

However, there are times when some targets are not covered by any camera, and we can't get their location even if we use centralized information. For example, we only know their location in time step $t$. In this case, we have no idea of the target's moving speed and direction, so we consider that the position of the target in time step $t+t_0$ is the same as that in time step $t$.
%\begin{equation}\label{positiondefault}
%\left\{
%\begin{aligned}
%\tilde{x}^p_{t+t_0} & = x^p_t \\
%\tilde{y}^p_{t+t_0} & = y^p_t
%\end{aligned}
%\right.
%.
%\end{equation}
Similarly, if we only know the position of time step $t-1$, we also can directly use it as the position estimation of time step $t+t_0$. If a target is not covered at any time in history, we will not predict the location of this target. The specific process can refer to Figure \ref{pre}. Since we can observe the positions of target 2, 4, and 5 in time step $t-1$ and $t$, then we estimate their positions use Equation \ref{positionpredict}, and we can only observe the positions of target 1, 6, and 8 in time step either $t-1$ or $t$, so we estimate their positions use their historical positions. However, target 3 and 7 can not be observed in any time step, we don't give their positions in the future prediction.

\subsection{Multi-agent Reinforcement Learning}

Considering each camera as an agent, our environment can be formulated as a partially observable Markov decision process (POMDP) model, which is defined as $(\mathcal{S},\mathcal{O},\mathcal{A},R,P,Z,\gamma)$, where $\mathcal{S},\mathcal{O},\mathcal{A},R,P,Z,\gamma$ represent the environment state space, observation space, action space, reward function, probability transfer function of the states, observation function, and discount factor, respectively.

In each time step $t$, the $i$-th camera obtains its observation $\boldsymbol{o}_{t,i} \in \mathcal{O}$ from the global environment state $\boldsymbol{s}_t \in \mathcal{S}$ with the probability $Z(\boldsymbol{o}_{t,i}|\boldsymbol{s}_{t})$. The joint action $\boldsymbol{a}_t \in \mathcal{A}$ is drawn from the policy $\pi(\boldsymbol{a}_t|\boldsymbol{o}_{t})$. After agents take the joint action $\boldsymbol{a}_t$, the environment receives a reward $r_t=R(\boldsymbol{s}_t,\boldsymbol{a}_t)$, which includes the individual reward for each agent and the team reward. Meanwhile, the environment state updates according to the probability transfer function $P(\boldsymbol{s}_{t+1}|\boldsymbol{s}_t,\boldsymbol{a}_t)$. Our goal is to maximize the expected accumulated reward $\mathbb{E}[\sum_{t=1}^{T}{\gamma^t r_t}]$.

\paragraph{Reward Function}
The reward function describes the goal of reinforcement learning. An effective reward function is very important for network learning. In this work, we divide the reward into team reward $R^t$ and individual reward $R^p_i$. Team reward describes the common goal of all cameras, that is, target coverage. We use $I_{ij}$ to indicate whether the $j$-th target is in the field of view of the $i$-th camera. If yes, $I_{ij}=1$, otherwise, $I_{ij}=0$. $n$ and $m$ represent the number of cameras and targets in the environment, respectively. Then, the team reward can be written as,
\begin{equation}\label{teamreward}
R^t = \frac{1}{m}\sum_{j=1}^{m}{\min{\{1, \sum_{i=1}^{n}{I_{ij}}\}}}.
\end{equation}
Different from team reward, individual reward pays more attention to the contribution of each camera to the target coverage. In order to maximize the overall target coverage, we expect each camera to cover different targets. Therefore, we take the ratio of the target covered by a single camera to the total number of targets as individual reward, and the targets covered by other cameras are not included. The $i$-th camera's individual reward can be written as,
\begin{equation}\label{individualreward}
R^p_i = \frac{1}{m}\sum_{j=1}^{m}{I_{ij}\max{\{0, 2-\sum_{i=1}^{n}{I_{ij}}\}}}.
\end{equation}
Finally, the total reward $R_i$ consists of team reward and every camera's individual reward, so it can be written as,
\begin{equation}\label{totalreward}
R_i = \lambda R^t + (1-\lambda) R^p_i.
\end{equation}

\begin{figure}[t]
\centering
\includegraphics[width=1.\linewidth]{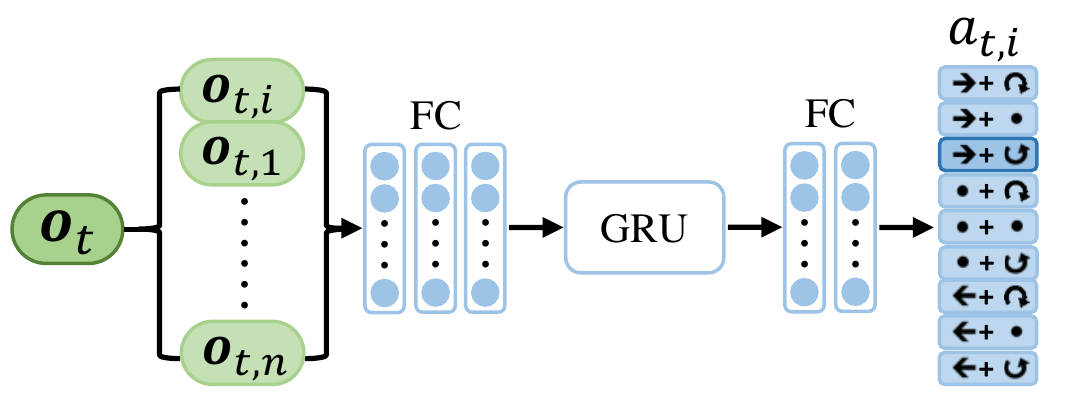}
\caption{The network structure of the $i$-th agent. The input of the network is the centralized observation $\boldsymbol{o}_{t}$, where the observation of the $i$-th camera $\boldsymbol{o}_{t,i}$ is ranked first, followed by others. The output of the network is the action of the $i$-th camera $a_{t,i}$.}
\label{figmarl}
\end{figure}

\paragraph{Network Structure}
We use the Q-Learning algorithm to train our model \cite{samvelyan19smac}. In our setting, the network includes 5 fully connected layers (FC) and a gated recurrent unit (GRU) which is shown in Figure \ref{figmarl}. All agents share network parameters, and we adjust the order of the observation to indicate different agent action from the network. For example, the $i$-th agent action $a_{t,i}$ in time step $t$ is sampled from $\pi(\cdot|\{\boldsymbol{o}_{t,i},\boldsymbol{o}_{t,1},\cdot,\boldsymbol{o}_{t,i-1},\boldsymbol{o}_{t,i+1},\cdots,\boldsymbol{o}_{t,n}\})$. %Finally, the action from MARL network can be written as $\boldsymbol{a}^{net}_t = \boldsymbol{a}_t = (a_{t,1}, a_{t,2}, \cdots, a_{t,n})$.

\subsection{Monte Carlo Tree Search}
%In the previous steps, we have estimated the future environment state $\tilde{\boldsymbol{s}}_{t+1},\tilde{\boldsymbol{s}}_{t+2},\cdots,\tilde{\boldsymbol{s}}_{t+D}$. Then, we can search the final action $\boldsymbol{a}^{mcts}_{t}$ based on this information. 

According to the setting of our 2D environment, each camera has 9 discrete action options. In a time step, the number of joint actions of $n$ cameras in the environment has reached $9^n$. If we search all actions, the branches of the search tree will be very large, which also leads to low search efficiency. To this end, we first prune the search tree using the action of multi-agent reinforcement learning network $\boldsymbol{a}^{net}_{t}$. We assume that the optimal action $\boldsymbol{a}^{mcts}_{t}$ only has one camera's action different from $\boldsymbol{a}^{net}_{t}$. Based on this assumption, the number of child nodes of the search tree reduces to $8n+1$. This search space is acceptable, and our experiments show that we can find a better action under this assumption.

Similar to the traditional MCTS method, each search step including selection, expansion, simulation, and backpropagation, and tree node in the search tree stores two vectors $N(s,a)$ and $V(s,a)$, which represent the number of visit times and action value, respectively. In most case, when a tree node is created, $N(s,a)$ and $V(s,a)$ will be initialized to 0, but in our method, we have $\boldsymbol{a}^{net}_{t}$ as a priori knowledge. Thus we initialize $N(s,a)$ as follows,
\begin{equation}\label{inite}
\left\{
\begin{aligned}
V(s, a_{ij}) & = \frac{q_{ij}}{\max \limits_{k}{q_{ik}}} \\
V(s, a_0)    & = 1
\end{aligned}
\right.
,
\end{equation}
where $a_0$ indicates the action from MARL network, $a_{ij}$ denotes the action of the $i$-th camera is changed to $j$, and $q_{ij}$ represents the q-value of the $i$-th camera with action $j$, which can be obtained from the output of the MARL network.

When a node is expanded, we use the MARL network rollout to the depth $D$. Since the prediction of the target location is only available for a short time period, the value of $D$ should not be too large. After simulation, in our task, we will generate rewards $r_i$ in each step of the exploration process. When updating the action value of a node, we are concerned about the average value of rewards of all nodes after this node. We assume that after the node to be updated, the reward sequence generated by our search is $r_{i+1}, r_{i+2}, \cdots, r_{i+k}$, then the average reward $r$ can be written as,
\begin{equation}\label{reup}
r = \frac{1}{k}\sum_{j=1}^{k}{r_{i+j}}.
\end{equation}
Then the $N(s,a)$ and $V(s,a)$ are updated as follows, 
\begin{equation}\label{update}
\left\{
\begin{aligned}
N(s, a) & \leftarrow N(s, a) + 1 \\
V(s, a) & \leftarrow V(s, a) + \frac{r - V(s, a)}{N(s, a) + 1}
\end{aligned}
\right.
.
\end{equation}
After $T$ search steps, we select the action with largest $N(s, a)$ as the cameras final action,
\begin{equation}\label{lastaction}
\boldsymbol{a}^{mcts}_{t} = \mathop{\arg\max} \limits_{a} N(s, a).
\end{equation}

\section{Experiments}
%In this section, we first introduce the 2D environment and then show the experiment results in detail.

\subsection{Experiment Setup}
\paragraph{Environment Setting}
As shown in Figure \ref{env}, in our 2D environment, the field of view of each camera can be represented by a sector centered on the camera. In our experiment, the visual angle (\emph{i.e.,} the center angle of the sector) and visual distance (\emph{i.e.,} the radius of the sector) are set as 800 and $90^{\circ}$, respectively. In each time step, the unit step of camera movement is 10 and the unit angle of rotation is $5^{\circ}$. In order to explore the influence of the number of targets and cameras and the size of the sports field on the experimental results, we set 6 different environments according to the parameters shown in Table \ref{envset}.

\begin{table}[htb]
\centering
\caption{Parameter of different environments. $n$ refers to the number of cameras, $m$ refers to the number of targets.}
\begin{tabular}{cp{0.8cm}<{\centering}p{0.8cm}<{\centering}c}
\hline
env name & $n$ & $m$ & sports field size \\
\hline
$Volleyball\_A$ & 6 & 12 & 2400 $\times$ 1200  \\
$Basketball\_A$ & 6 & 10 & 2240 $\times$ 1200 \\
$Football\_A$   & 6 & 22 & 2100 $\times$ 1360  \\
$Volleyball\_B$ & 4 & 12 & 2400 $\times$ 1200  \\
$Basketball\_B$ & 4 & 10 & 2240 $\times$ 1200 \\
$Football\_B$   & 4 & 22 & 2100 $\times$ 1360  \\
\hline
\end{tabular}
\label{envset}
\end{table}

\paragraph{Target Movement}
In our environment, all targets have their moving goals and they move toward the goals with random speed between $v$ and $1.2v$. $v$ is sampled from a uniform distribution. In our experiment, the upper and lower bound of the uniform distribution are set to 50 and 100, respectively. When a target is close to the goal, the environment will generate a new goal for the target.

\paragraph{Environment Initialization}
We have tried to initialize the camera location in three different ways: (1) Random: cameras randomly initialize around the sports fields. (2) Part: divide the sports fields edge into $n$ segments, and each camera is randomly initialized in each segment. (3) Fix: cameras initialize at $n$ fixed points. In the first method, cameras are likely to be initialized in close locations. In order to cover more targets, cameras will spend much time moving away from each other. However, for the second method, if there are near parts between segments, it can not completely solve the above problem. Through experiments in three environments, we find that the third method can obtain the highest target coverage. The experiment result is shown in Table \ref{init}. In the following experiments, we all use the third method to initialize the camera location, and the visual angle of cameras and the position of the target we are randomly initialized.

\begin{table}[htb]
\centering
\caption{Target coverage ($\%$) results with different camera initialization methods.}
\begin{tabular}{cccc}
\hline
env name & random & part & fix \\
\hline
$Volleyball\_A$ & 77.8 $\pm$ 6.9 & 83.0 $\pm$ 6.4 & 86.1 $\pm$ 4.1  \\
$Basketball\_A$ & 71.8 $\pm$ 11.3 & 82.5 $\pm$ 9.4 & 85.7 $\pm$ 4.8 \\
$Football\_A$   & 67.7 $\pm$ 11.6 & 79.6 $\pm$ 5.4 & 86.6 $\pm$ 3.4  \\
\hline
\end{tabular}
\label{init}
\end{table}

\begin{table*}[t]
\centering
\caption{Target coverage ($\%$) result in different environments. We demonstrate the impact of each module on the experimental results through ablation study, and compare our results with HiT-MAC. The best result in each environment is shown in bold.}
\begin{tabular}{cp{2.1cm}<{\centering}p{2.1cm}<{\centering}p{2.17cm}<{\centering}p{2.1cm}<{\centering}p{2.1cm}<{\centering}p{2.1cm}<{\centering}}
\hline
 & random & MARL action & MARL random & Ours- & Ours & HiT-MAC\\
\hline
 MARL & $\times$ & $\checkmark$ & $\checkmark$ & $\checkmark$ & $\checkmark$ & $\checkmark$\\
 prediction & $\times$ & $\times$ & $\times$ & $\times$ & $\checkmark$ & $\times$ \\
 MCTS & $\times$ & $\times$ & $\times$ & $\checkmark$ & $\checkmark$ & $\times$ \\
\hline
$Volleyball\_A$ & 40.4 $\pm$ 19.1 & 86.1 $\pm$ 4.1 & 85.3 $\pm$ 4.3 & 88.5 $\pm$ 3.9 & $\textbf{90.4}$ $\mathbf{\pm}$ $\textbf{4.1}$ & 76.2 $\pm$ 13.0 \\
$Basketball\_A$ & 34.6 $\pm$ 21.4 & 85.7 $\pm$ 4.8 & 84.1 $\pm$ 5.4 & 88.6 $\pm$ 4.3 & $\textbf{89.7}$ $\mathbf{\pm}$ $\textbf{4.1}$ & 79.1 $\pm$ 13.6 \\
$Football\_A$   & 33.6 $\pm$ 17.0 & 86.6 $\pm$ 3.4 & 84.4 $\pm$ 4.4 & 89.0 $\pm$ 3.2 & $\textbf{90.7}$ $\mathbf{\pm}$ $\textbf{3.1}$ & 84.9 $\pm$ 6.5 \\
$Volleyball\_B$ & 11.3 $\pm$ 7.8 & 56.9 $\pm$ 12.9 & 52.5 $\pm$ 12.4 & 59.1 $\pm$ 13.5 & $\textbf{59.9}$ $\mathbf{\pm}$ $\textbf{13.7}$ & 51.6 $\pm$ 7.2 \\
$Basketball\_B$ & 12.0 $\pm$ 9.2 & 60.8 $\pm$ 12.0 & 56.4 $\pm$ 11.1 & 62.8 $\pm$ 11.6 & $\textbf{64.3}$ $\mathbf{\pm}$ $\textbf{12.7}$ & 50.4 $\pm$ 10.0 \\
$Football\_B$   & 14.1 $\pm$ 10.6 & 61.5 $\pm$ 8.3 & 57.1 $\pm$ 9.9 & 64.4 $\pm$ 8.5 & $\textbf{65.5}$ $\mathbf{\pm}$ $\textbf{9.1}$ & 54.9 $\pm$ 6.1 \\
\hline
\end{tabular}
\label{result}
\end{table*}

\paragraph{Hyper Parameters}
For the learning of the multi-agent reinforcement learning network, the learning rate $\eta$, discount factor $\gamma$, team reward weight $\lambda$ are 0.0005, 0.99, and 0.1, respectively. For the Monte Carlo tree search, the search depth $D$, simulation steps $T$ are 3 and 100, respectively. We train the MARL model and test the environment with NVIDIA GeForce RTX 2080Ti GPU.

\begin{figure}[thb]
\centering
\includegraphics[width=0.98\linewidth]{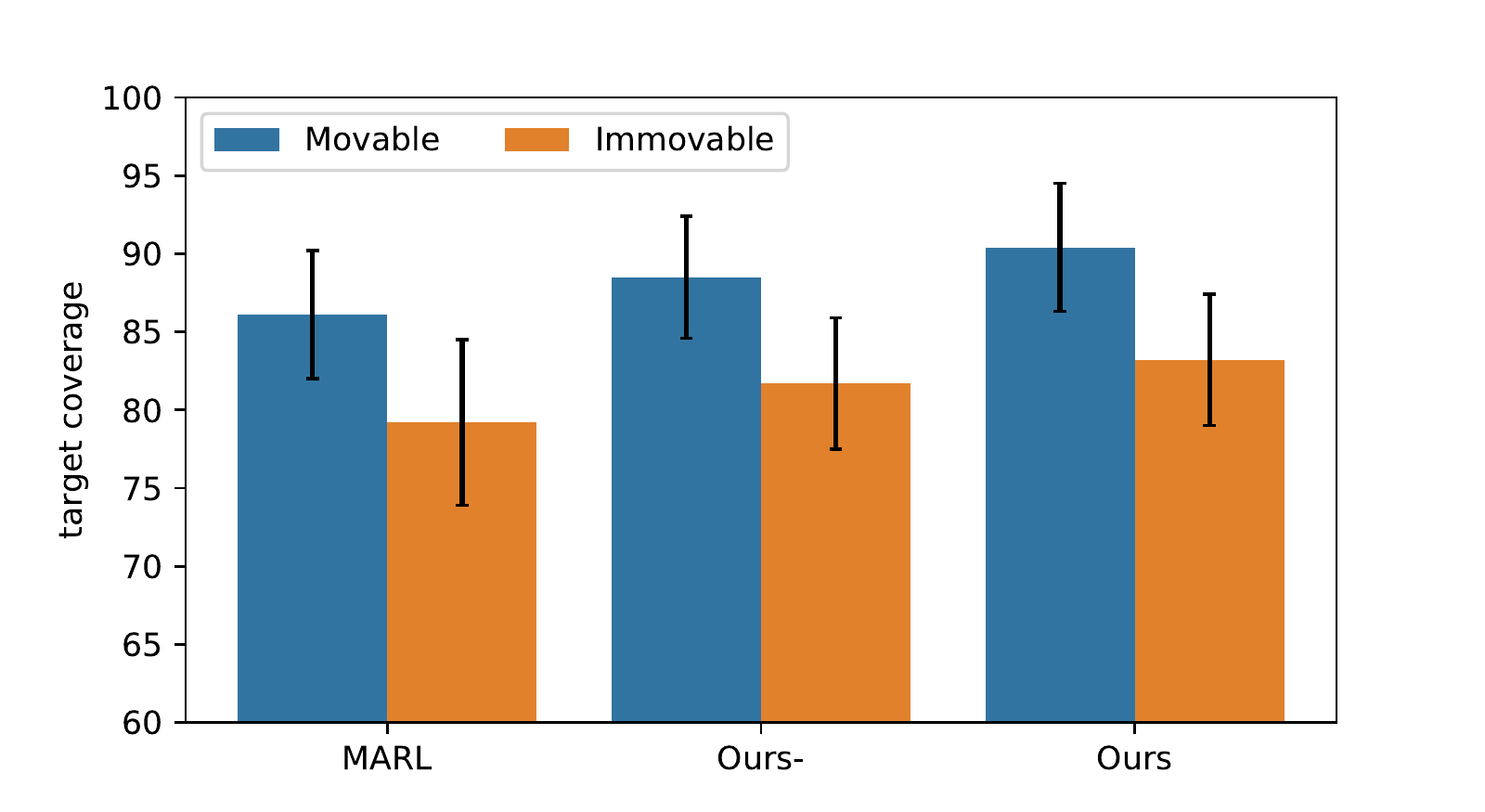}
\caption{Contribution of camera's degree of freedom in position to target coverage ($\%$) in $Volleyball\_A$ environment.}
\label{immove}
\end{figure}

\subsection{Experimental Results}
In order to prove the effectiveness of our method, we conduct corresponding ablation experiments in each environment. Table \ref{result} shows the performance of the model under different settings. Starting from the second column of the Table \ref{result}, ``random'' indicates the action is random choose from the action space, without learning at all. ``MARL action'' indicates we use the output of the MARL network $\boldsymbol{a}^{net}_t$ as the final action, without position prediction and MCTS. In our MCTS method, we assume that the optimal action only has one camera's action different from $\boldsymbol{a}^{net}_{t}$, so we try to find the best action in $8n+1$ actions. ``MARL random'' indicates after we get the $\boldsymbol{a}^{net}_t$ from MARL network, we randomly choose an action as final action from $8n+1$ actions without MCTS. ``Ours-'' removes the position prediction module, we assume the future environment state is same as current environment state, then use MCTS to find the final action. ``Ours'' includes all modules: MARL, position prediction, and MCTS. The last column shows the performance of HiT-MAC \cite{xu2020learning}. According to the Table \ref{result}, we can see our method performs best and each module contributes to the target coverage. 

\begin{figure}[thb]
\centering
\includegraphics[width=0.98\linewidth]{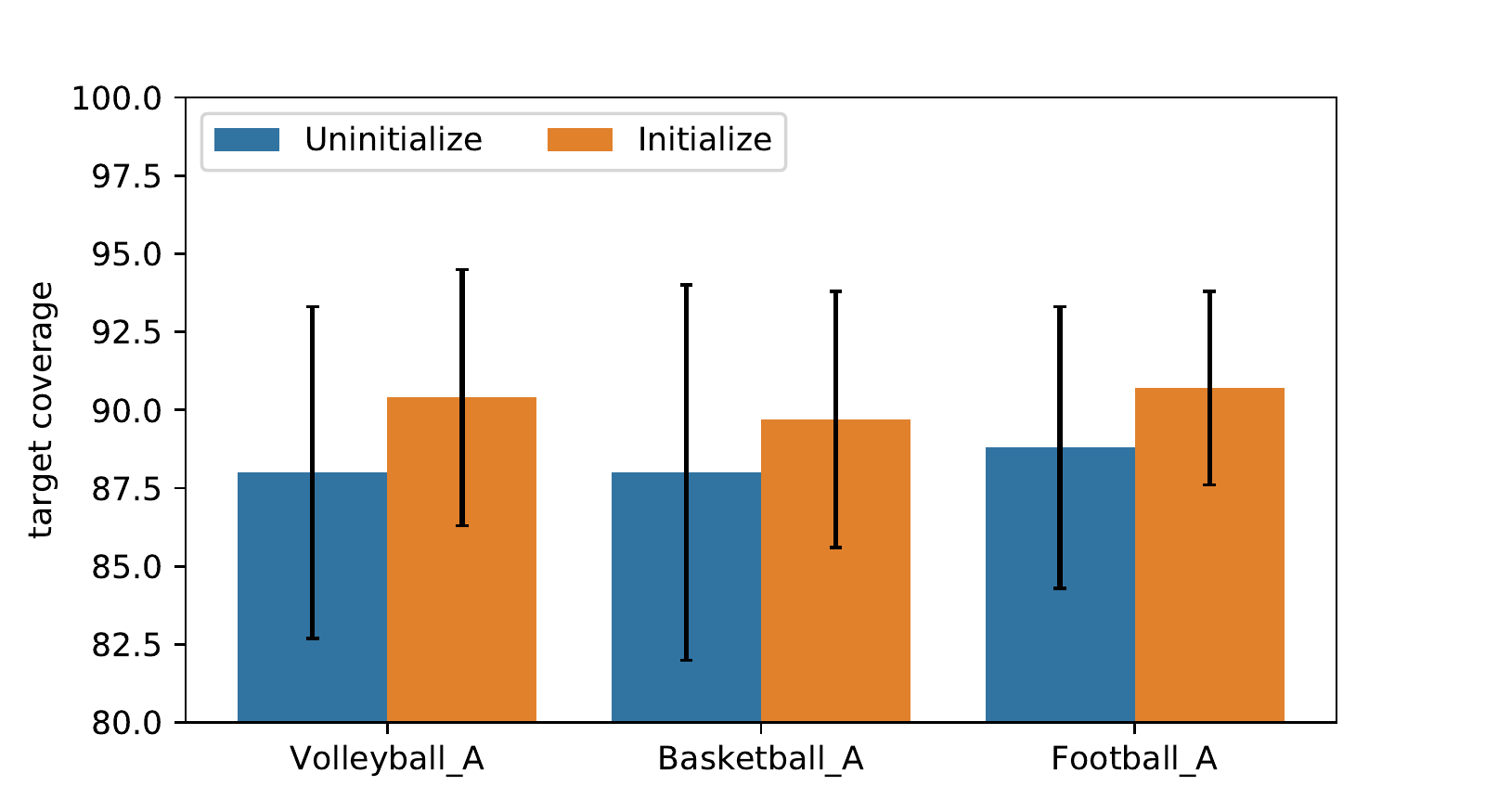}
\caption{Performance improvement of target coverage ($\%$) by the initialization of $V(s,a)$ in MCTS.}
\label{uninit}
\end{figure}

\begin{figure}[!h]
\centering
\includegraphics[width=0.98\linewidth]{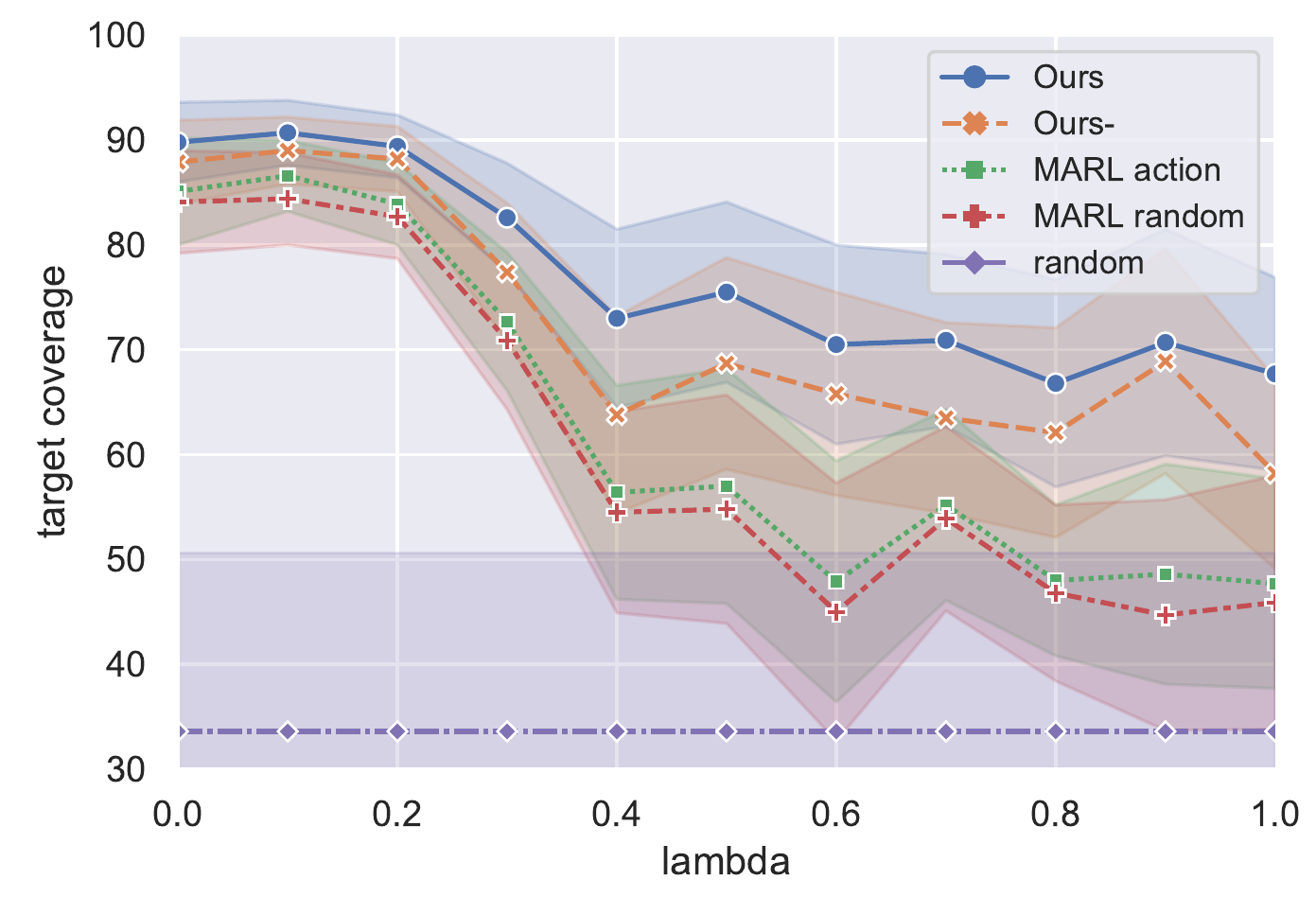}
\caption{The impact of team reward weight $\lambda$ on target coverage ($\%$) in $Football\_A$ environment.}
\label{lambda}
\end{figure}

In addition, we compare our method with the case where the camera cannot move in location. As shown in Figure \ref{immove}, the camera has higher degrees of freedom achieves higher target coverage. And Figure \ref{uninit} shows the initialization of $V(s,a)$ in MCTS can effectively improve the target coverage. Figure \ref{lambda} shows impact of different team reward weight $\lambda$ on target coverage in the $Football\_A$ environment.

\section{Conclusion}
In this work, we propose a novel multi-target active tracking method with MCTS. Based on the prediction of the future position for the targets, we obtain the optimal action through MCTS method. Meanwhile, we use the action from MARL network to prune the search tree, which greatly reduced the search space of MCTS method. In future work, we will use a more accurate model to estimate the target position, which makes the search results more reliable. %What's more, we will replace 2D coordinates with bounding box, and realize multi-target coverage in 3D environment combined with visual images.

\bibliographystyle{named}
\bibliography{ijcai22}

\end{document}